\def\paperID{12935}
\def\confName{CVPR}
\def\confYear{2026}

\def\paperTitle{RE-VLM: Event-Augmented Vision-Language Model for Scene Understanding}

\def\authorBlock{

Hanqing Liu$^{1}$\thanks{Equal contribution.}, Mingjie Liu$^{1}$\footnotemark[1], Luoping Cui$^{1}$, Endian Lin$^{1}$, Donghong Jiang$^{1}$, Chuang Zhu$^{1,2}$\thanks{Corresponding author.}\\

$^{1}$School of Artificial Intelligence, Beijing University of Posts and Telecommunications, Beijing, China\\
$^{2}$State Key Laboratory of General Artificial Intelligence, BIGAI, Beijing, China\\

\texttt{\{hanqingliu, LMJ, lpcui, ledgogo7, donghongjiang, czhu\}@bupt.edu.cn}

}
\newif\ifreview 
\newif\ifarxiv \newcommand{\arxiv}{\arxivtrue}
\newif\ifcamera 
\newif\ifrebuttal 

\arxiv
\pdfoutput=1
\documentclass[10pt,twocolumn,letterpaper]{article}
\ifreview \usepackage[review]{cvpr} \fi
\ifarxiv \usepackage[pagenumbers]{cvpr} \fi
\ifrebuttal \usepackage[rebuttal]{cvpr} \fi
\ifcamera \usepackage{cvpr} \fi

\usepackage{graphicx}	
\usepackage{amsmath}	
\usepackage{amssymb}	
\usepackage{booktabs}
\usepackage{times}
\usepackage{microtype}
\usepackage{epsfig}
\usepackage{caption}
\usepackage{float}
\usepackage{placeins}
\usepackage{color, colortbl}
\usepackage{stfloats}
\usepackage{enumitem}
\usepackage{tabularx}
\usepackage{xstring}
\usepackage{multirow}
\usepackage{xspace}
\usepackage{url}
\usepackage{subcaption}
\usepackage{xcolor}
\usepackage[table]{xcolor}
\usepackage{pifont}
\usepackage{booktabs}
\usepackage{multirow}
\usepackage{xcolor}
\newcommand{\cmark}{\ding{51}}
\newcommand{\xmark}{\ding{55}}
\usepackage[hang,flushmargin]{footmisc}

\ifcamera \usepackage[accsupp]{axessibility} \fi

\ifarxiv  \fi

\newcommand{\R}[1]{{%
    \textbf{%
        \ifstrequal{#1}{1}{\textcolor{red}{R#1}}{%
        \ifstrequal{#1}{2}{\textcolor{blue}{R#1}}{%
        \ifstrequal{#1}{3}{\textcolor{magenta}{R#1}}{%
        \ifstrequal{#1}{4}{\textcolor{teal}{R#1}}{%
                           \textcolor{cyan}{R#1}%
        }}}}%
    }%
}}

\usepackage{xr-hyper}

\makeatletter
\newcommand*{\addFileDependency}[1]{
  \typeout{(#1)}
  \@addtofilelist{#1}
  \IfFileExists{#1}{}{\typeout{No file #1.}}
}

\makeatother
\newcommand*{\myexternaldocument}[1]{
    \externaldocument{#1}
    \addFileDependency{#1.tex}
    \addFileDependency{#1.aux}
}

\definecolor{cvprblue}{rgb}{0.21,0.49,0.74}
\usepackage[pagebackref,breaklinks,colorlinks,allcolors=cvprblue]{hyperref}
\usepackage[capitalize]{cleveref}
\crefname{section}{Sec.}{Secs.}
\crefname{table}{Table}{Tables}
\crefname{figure}{Fig.}{Figs.}

\ifarxiv \crefname{appendix}{App.}{Apps.}
\else \crefname{appendix}{Suppl.}{Suppls.} \fi

\unless\ifarxiv \myexternaldocument{_supplementary} \fi

\usepackage{arydshln}
\usepackage{booktabs}
\begin{document}

\title{\paperTitle}
\author{\authorBlock}
\maketitle 

\begin{abstract}
Conventional vision-language models (VLMs) struggle to interpret scenes captured under adverse conditions (e.g., low light, high dynamic range, or fast motion) because standard RGB images degrade in such environments. Event cameras provide a complementary modality: they asynchronously record per-pixel brightness changes with high temporal resolution and wide dynamic range, preserving motion cues where frames fail. We propose RE-VLM, the first dual-stream vision-language model that jointly leverages RGB images and event streams for robust scene understanding across both normal and challenging conditions. RE-VLM employs parallel RGB and event encoders together with a progressive training strategy that aligns heterogeneous visual features with language. To address the scarcity of RGB-Event-Text supervision, we further propose a graph-driven pipeline that converts synchronized RGB-Event streams into verifiable scene graphs, from which we synthesize captions and question--answer (QA) pairs. To develop and evaluate RE-VLM, we construct two datasets: PEOD-Chat, targeting illumination-challenged scenes, and RGBE-Chat, covering diverse scenarios. On captioning and VQA benchmarks, RE-VLM consistently outperforms state-of-the-art RGB-only and event-only models with comparable parameter counts, with particularly large gains under challenging conditions. These results demonstrate the effectiveness of event-augmented VLMs in achieving robust vision-language understanding across a wide range of real-world environments. 
\end{abstract}

\section{Introduction}
Vision-language models (VLMs) have made remarkable progress on tasks such as image captioning and visual question answering (VQA), demonstrating a strong ability to interpret and describe visual scenes~\cite{bai2025qwen2,chen2024internvl,lu2024deepseek}. However, most existing VLMs rely on high-quality RGB images and struggle under adverse conditions such as extreme low light or fast motion. In low-light scenes, overexposed or underexposed frames, or high-speed scenarios with significant motion blur, even state-of-the-art RGB-only VLMs often fail to produce accurate and meaningful descriptions. This limitation significantly constrains their applicability in real-world dynamic environments. 
\begin{figure}[t]
    \centering
    \includegraphics[trim=0.9cm 4cm 14cm 0.5cm, clip, width=0.45\textwidth]{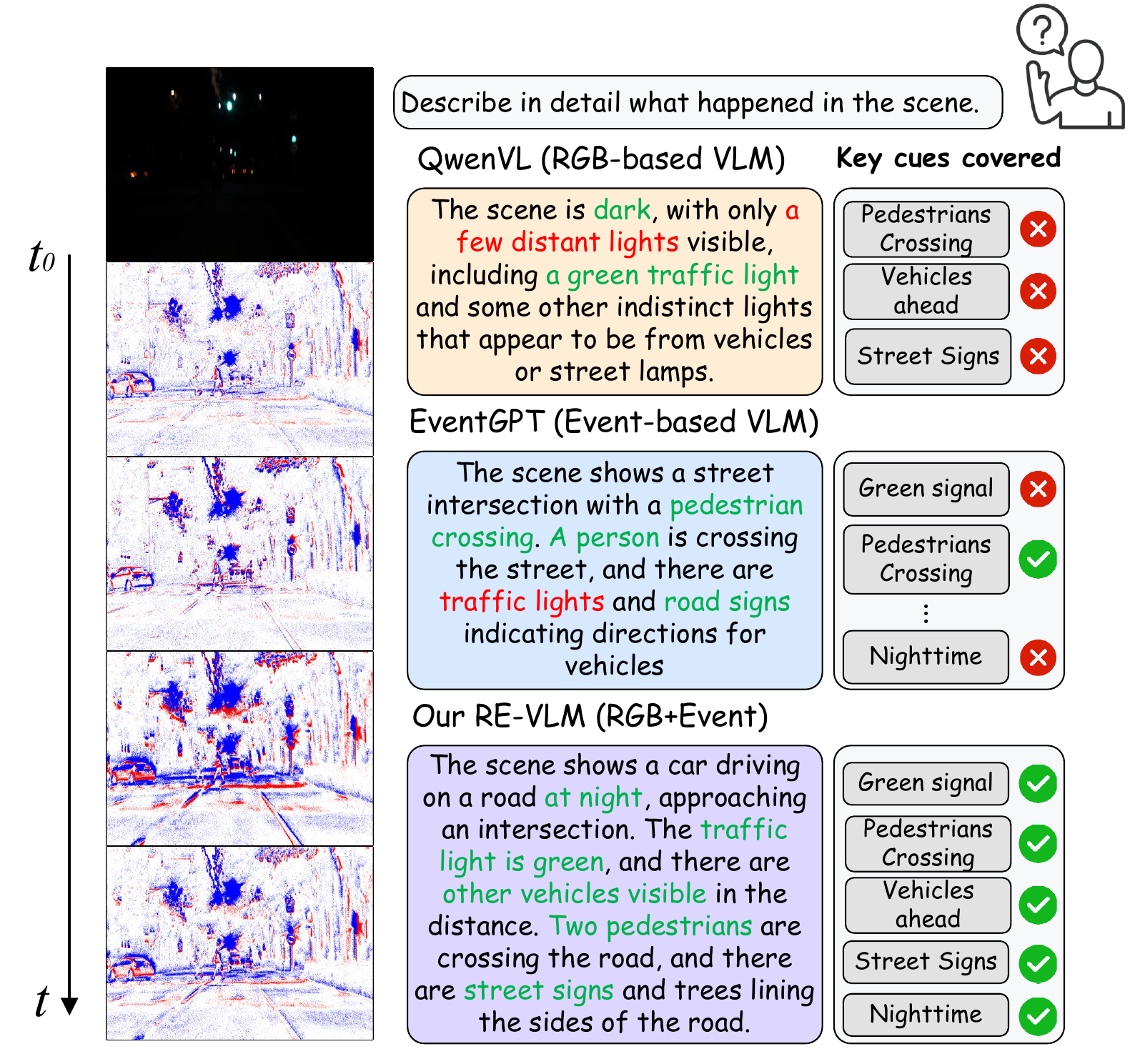}
    \caption{Illustration of RGB-Event complementarity in a challenging low-light scene. \textbf{RGB-only VLM:} Struggles with low dynamic range, missing the pedestrian and crosswalk. \textbf{Event-only VLM:} Captures the scene structure and the pedestrian's motion but lacks color and texture, failing to determine the traffic light's state. \textbf{Our RE-VLM:} Provides a complete description, correctly identifying the green light, the pedestrian, the crosswalk, and other vehicles.
}
    \label{fig:introduction}
\end{figure}

In contrast to standard frame-based cameras, event cameras provide a complementary modality. Instead of capturing full image frames at fixed rates, an event camera asynchronously records per-pixel brightness changes with microsecond-level latency and high dynamic range~\cite{gallego2020event}. This unique mechanism preserves fast motion cues while avoiding saturation and motion blur, enabling the capture of subtle scene changes even in low-light or high-contrast scenarios~\cite{vidal2018ultimate}. These properties make event data especially useful in conditions that plague conventional cameras. Recent multimodal vision-language efforts based on event streams (e.g., EventGPT~\cite{liu2025eventgpt}) have begun to leverage this modality; however, event streams alone lack rich appearance information (e.g., color and fine texture)~\cite{zhou2022rgb} essential for comprehensive scene understanding. As illustrated in \cref{fig:introduction}, the inherent complementarity between RGB frames and event streams motivates a unified RGB-Event vision-language framework that can capitalize on the strengths of both.

\begin{figure*}[t]
    \centering
    \includegraphics[trim=25cm 20cm 21.5cm 18cm, clip, width=1\textwidth]{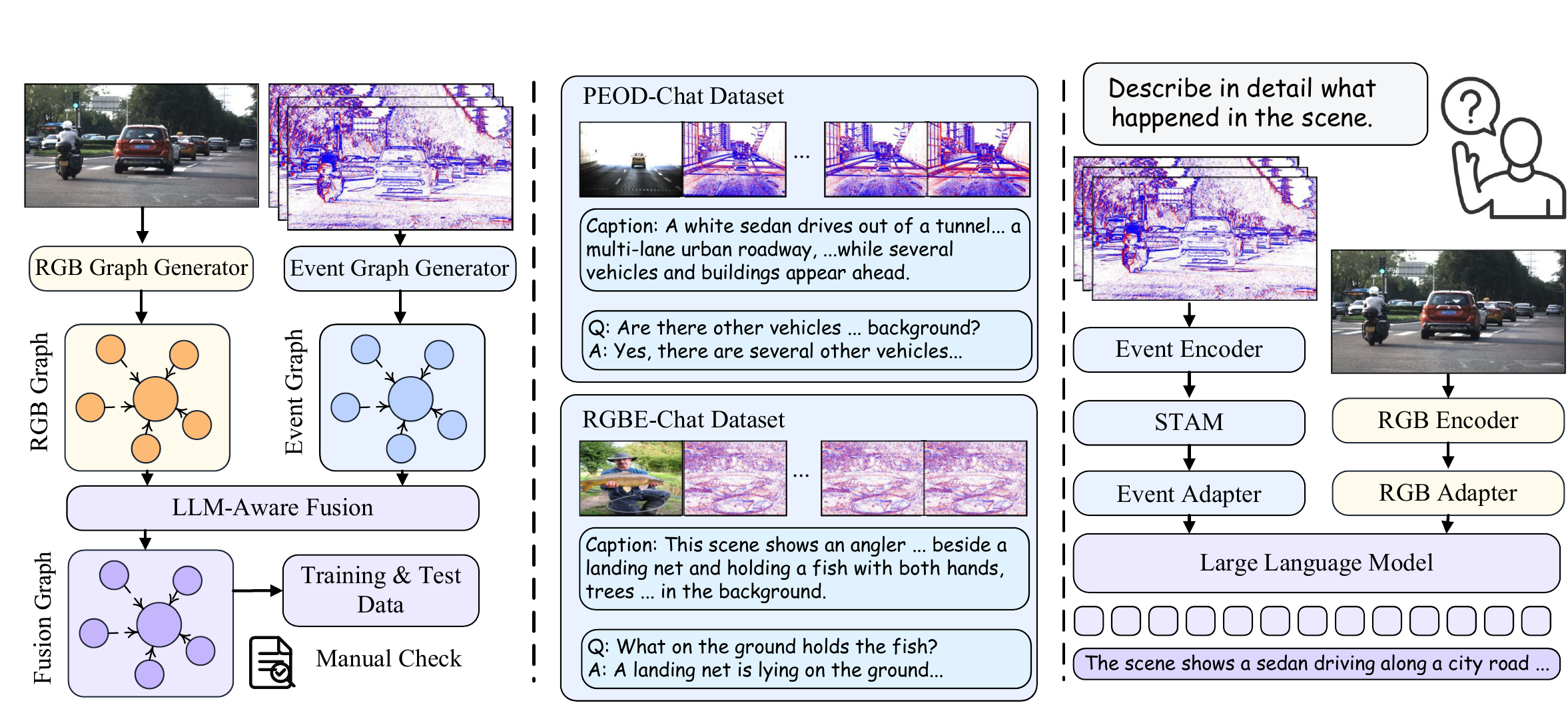}
    \caption{\textbf{Construction of RE-VLM: data generation pipeline and model.} \textbf{Left:} A graph-driven pipeline converts synchronized RGB frames and event streams into a graph, extracts verifiable scene facts, and synthesizes reliable caption and QA supervision.
    \textbf{Center:} Representative examples from the datasets yielded by the pipeline: \textbf{PEOD-Chat} (illumination-challenged scenes) and \textbf{RGBE-Chat} (general scenarios).
    \textbf{Right:} The RE-VLM model fuses static RGB appearance features with dynamic cues from events and is trained with a progressive strategy for robust captioning and VQA under low-light, high-dynamic-range, and fast-motion conditions.}
    \label{fig:overview}
\end{figure*}

To address these limitations, we present \textbf{RE-VLM}, the first VLM to fuse static RGB appearance cues (texture/color) with dynamic event-based cues (motion/high dynamic range). RE-VLM is a dual-stream model that also supports single-modality inputs and is designed for robust scene understanding under both standard and adverse conditions. A key challenge in developing such a model is the scarcity of RGB-Event-Text training data. We tackle this data bottleneck with a graph-driven data generation pipeline: synchronized RGB-Event streams are converted into an explicit RGB-Event knowledge graph capturing verifiable scene facts, and from this graph we automatically synthesize high-quality image captions and VQA pairs at scale. \cref{fig:overview} provides an overview of our approach. The left panel illustrates the graph-driven pipeline that converts synchronized RGB-Event streams into an explicit knowledge graph and provides human-verified captions and VQA annotations, yielding two new datasets: PEOD-Chat (focusing on challenging illumination scenarios) and RGBE-Chat (covering general scenes). The right panel depicts the RE-VLM architecture, which employs a dual visual encoder (RGB and event) with an LLM-aware fusion module. We detail the data pipeline in \cref{sec:data_engineering} and the model in \cref{sec:method}. Our main contributions are as follows:
\begin{itemize}
    \item \textbf{A robust RGB-Event dual-stream VLM.} We propose \textbf{RE-VLM}, the first VLM to fuse static RGB cues (texture/appearance) with dynamic event cues (motion/HDR). We introduce a progressive training regimen to effectively align these modalities, significantly improving robustness in low light, HDR transitions, and fast motion.
    \item \textbf{Graph-driven data pipeline and new RGB-Event-Text datasets.} To address the lack of RGB-Event-Text training data, we develop a graph-driven data generation pipeline. This pipeline synthesizes verifiable scene descriptions and QA pairs from synchronized RGB-Event inputs. Using this approach, we construct two datasets: PEOD-Chat (with challenging illumination conditions) and RGBE-Chat (covering general scenarios).
    \item \textbf{Impressive performance in RGB-Event understanding.} RE-VLM outperforms strong RGB-only and event-only baselines, including models with comparable or even $2\times$ more parameters, yielding notable gains in caption fidelity and situational understanding under adverse conditions.
\end{itemize}

\section{Related Work}
\paragraph{Vision-Language Models on RGB Images.} 
Large vision-language models (VLMs) have driven rapid progress in image understanding and multimodal reasoning. In academia and the open-source community, models like LLaVA~\cite{liu2023visual}, InternVL~\cite{chen2024internvl}, DeepSeek-VL~\cite{lu2024deepseek}, and Qwen2.5-VL~\cite{bai2025qwen2} integrate powerful vision encoders with large language models, achieving state-of-the-art results on diverse visual-linguistic tasks (e.g., image captioning and VQA). Meanwhile, industry-scale systems such as GPT-4V~\cite{hurst2024gpt} and Gemini~\cite{team2023gemini} advance the frontier with unprecedented multimodal capabilities, demonstrating the power of coupling visual inputs with LLMs. These models share a common limitation: they rely on static RGB images and remain vulnerable under adverse imaging conditions, where RGB-only models can fail to produce accurate or reliable predictions.

\noindent\textbf{Event-based vision and Event-Language alignment.} In contrast to standard cameras, event cameras offer a fundamentally different sensing modality that is well suited for challenging conditions~\cite{gehrig2024low}. A rich body of work has explored event-based vision for recognition, tracking, and SLAM~\cite{gallego2020event}. Researchers have developed methods for low-level perception (feature detection, optical flow)~\cite{mueggler2017fast,gehrig2018asynchronous} and high-level tasks like object recognition and 3D reconstruction using events~\cite{liu2025beyond,kim2016real,rebecq2018emvs}. These works demonstrate that event streams can effectively complement or substitute RGB inputs in scenarios with fast dynamics or HDR lighting~\cite{vidal2018ultimate}.

Integrating event data into vision-language frameworks has emerged as a promising research direction. Initial efforts (e.g. EventCLIP~\cite{wu2023eventclip}) adapt pre-trained image--text models to the event domain, enabling zero-shot and few-shot recognition on event data by capitalizing on CLIP's vision--text alignment. Similarly, EventBind~\cite{zhou2024eventbind} extends CLIP~\cite{radford2021learning} with a dedicated event encoder to embed events, images, and text in one space, supporting cross-modal retrieval and open-set classification under limited supervision. However, both EventCLIP~\cite{wu2023eventclip} and EventBind~\cite{zhou2024eventbind} lack a large language model and cannot leverage the vast world knowledge and linguistic context that an LLM provides. This limits their capacity for higher-level scene understanding, free-form description, or dialogue about event data. 

\noindent\textbf{Vision-Language Models with Events.}
Very recent studies integrate event data directly into multimodal LLMs. EventGPT~\cite{liu2025eventgpt} is the first VLM specifically tailored for event streams and shows strong performance under extreme motion and lighting extremes. Nonetheless, event-only VLMs inevitably inherit the fundamental limitation of the event modality: because events only capture changes, they contain no explicit color information and only sparse details about static scene context~\cite{gallego2020event}. In practice, such models can describe motion and high-contrast structure, but cannot recognize fine appearance attributes (e.g., object colors and textures) that are readily apparent in RGB images. This lack of rich appearance cues constrains the breadth of semantic understanding such models can provide. Meanwhile, the complementary strengths of frames and events suggest that a hybrid approach could unlock robust vision-language understanding across all conditions~\cite{gehrig2024low}. Indeed, prior works in low-level vision have shown that fusing event streams with standard frames yields superior performance in object detection and tracking under difficult conditions~\cite{zhou2022rgb,liu2024enhancing,zhang2023frame,dong2021standard}; overall, research on event cameras is shifting from event-only modeling toward RGB-Event fusion and from low-level vision toward higher-level scene understanding.

\section{RGB-Event-Text Data Generation}
\label{sec:data_engineering}

RGB image datasets have been extensively explored in computer vision. However, large-scale RGB-Event-Text corpora suitable for training multimodal large language models (MLLMs) remain scarce, especially for adverse conditions where a single modality degrades (e.g., low light, HDR transitions, or fast motion). Prior work has shown that large models (e.g., GPT-family) can synthesize supervision for multimodal applications. Yet, current pipelines that generate Event-Text data from RGB alone struggle precisely when RGB deteriorates, because RGB-trained VLMs cannot reliably infer content under severe degradation. To address this gap, we introduce a \textbf{graph-driven degradation-adaptive} data generation pipeline. A graph serves as a verifiable intermediate representation and explicitly encodes degradation cues to guide modality weighting. The pipeline converts synchronized RGB and event streams into graph-structured scene facts and then synthesizes caption and VQA supervision under degraded conditions.

\begin{figure*}[t]
    \centering
    \includegraphics[trim=0.7cm 16cm 0.8cm 1.2cm, clip, width=1\textwidth]{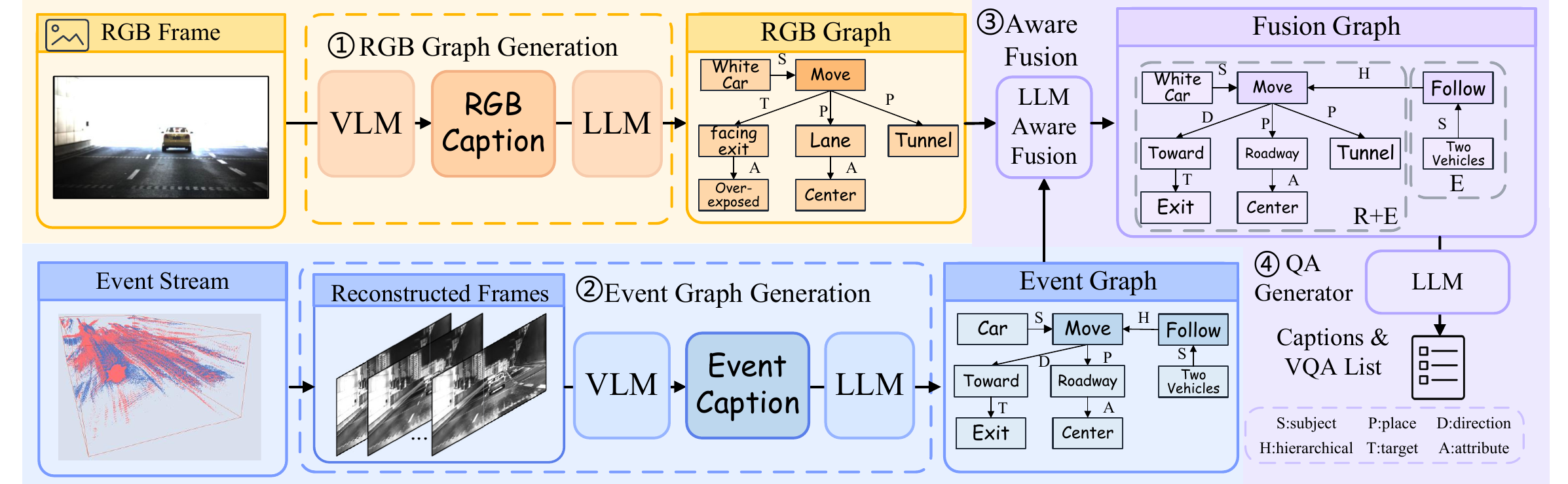}
    \caption{\textbf{Data generation pipeline overview.} From reconstructed event frames and RGB images, two modality-specific graphs are constructed. A degradation-aware fusion then merges them into a single RGB-event graph (nodes: entities, edges: relations). Finally, captions and VQA items are synthesized from the fused graph. (S: subject, P: place, D: direction, T: target; H: hierarchical relation; A: attribute.)}
    \label{fig:pipeline}
\end{figure*}
\subsection{Graph Generation}
\label{subsec:graph}

\paragraph{Event Graph Generation.}  To ensure temporal consistency with the image modality, we first locate each RGB keyframe and select an $N\times 33\,\mathrm{ms}$ event window, where $N=4$, centered on its timestamp. We reconstruct this event segment into a sequence of $N$ grayscale frames using a state-of-the-art event reconstruction network (e.g., NER-Net~\cite{liu2024seeing}). Stacking the reconstructed frames in temporal order yields a ``video-like'' event tensor that serves as input to a captioning VLM.

Following the ``Captioning'' step in \cref{fig:pipeline}, we generate a structured description constrained to observable facts. The description follows a subject--motion--place--relation schema, maintains consistent coreference, and explicitly marks dynamic and temporal cues. Entity qualifiers (e.g., relative position, appearance) are kept concrete and stable to facilitate downstream parsing.

In the subsequent ``Graph Parsing'' stage, we use an LLM to convert the caption into an event graph. Each node is labeled with a minimal argument tuple (subject, motion, place, attributes); for example,
$\mathrm{Move}(\text{subject}=\text{car},\,\text{motion}=\text{forward},\,\text{place}=\text{lane\_center})$.
Standardized naming ensures graph coreference consistency. The resulting graph is a concise, auditable intermediate representation that supports degradation-aware fusion and subsequent QA generation.

\noindent\textbf{RGB Graph Generation.}
To remain strictly time-aligned with the event stream, we build the RGB graph on the RGB keyframe synchronized with the selected event window. In contrast to the event graph's emphasis on temporal contours and motion trends, the RGB graph focuses on appearance and static structure, prioritizing entity attributes (color, texture, shape), scene geometry, and the global spatial layout. We also incorporate degradation phenomena into both the description and the structured graph by explicitly annotating degradations such as low light, overexposure, glare, motion blur, etc. These degradation labels are attached to the relevant nodes and edges and later serve as principled evidence for modality weighting in graph fusion.

\subsection{Graph Fusion and QA Generation}
We employ an LLM to perform degradation-aware fusion between the event graph and the RGB graph. Concretely, we propose a degradation-adaptive fusion strategy and denote the event graph by $G_e$ and the RGB graph by $G_r$. We diagnose imaging quality solely from $G_r$ based on its degradation labels. We then perform field-level arbitration during fusion: all facts involving motion cues, temporal ordering, and topological structure are anchored to $G_e$ because event signals are more robust to dynamics and edge structures; facts involving light sources, color, or readable text are taken from $G_r$, provided that $G_r$ is not severely degraded. If $G_r$ is degraded, its conclusions must not override those from $G_e$; instead, they are retained as low-confidence candidates.

 For geometry-related fields such as counts and positions, we adopt the consensus if $G_e$ and $G_r$ agree; otherwise, $G_r$ takes precedence and the outputs from $G_e$ are downgraded to secondary. We align and normalize entities, and attach metadata such as modality presence and confidence. Semantically equivalent events are merged, annotated with \texttt{source} $\in \{G_e, G_r, G_{e+r}\}$ and confidence scores. Finally, we feed the fused graph and fusion policy into a text-generation model to generate caption and up to three VQA items.

To assess the reliability of our generation pipeline, we adopt a human-audited correction protocol, following recent dataset-curation practice that performs large-scale caption correction and validates quality via human studies~\cite{luo2024view}. Concretely, we randomly sampled $N\!=\!855$ instances from PEOD and, for each instance, produced textual supervision (QA pairs) using two systems: (i) our method and (ii) an RGB-only generation baseline (EventGPT~\cite{liu2025eventgpt}). Human annotators audited every item and marked it as requiring correction if any factual error appeared in either the question or the answer. As summarized in \cref{tab:qa_correction_rate_peod}, our approach yields a substantially lower correction rate, indicating more reliable supervision in adverse scenes.

\begin{table}[hbtp]
\centering
\caption{Manual QA corrections on PEOD samples. Human-audited correction rate and count comparing an RGB-only generation baseline~\cite{liu2025eventgpt} with our method; lower is better.}
\small
\setlength{\tabcolsep}{8pt}
\renewcommand{\arraystretch}{1.1}
\begin{tabular}{lcc}
\toprule
\textbf{Method} & \textbf{Correction Rate} & \textbf{Correction Count} \\
\midrule
RGB VLM~\cite{liu2025eventgpt} & 54.2\% & 463 \\
Ours                     & 18.1\% & 155 \\
\bottomrule
\end{tabular}

\label{tab:qa_correction_rate_peod}
\end{table}

\subsection{PEOD-Chat and RGBE-Chat}In challenging scenarios such as the PEOD dataset~\cite{cui2025peodpixelalignedeventrgbbenchmark}, 
 we apply the proposed pipeline to synthesize caption and QA supervision from synchronized RGB-Event streams. After manual screening, the resulting items form \textbf{PEOD-Chat}, an RGB-Event-Text dataset for challenging illumination conditions. For more general scenarios, we follow the dataset construction protocol of EventGPT~\cite{liu2025eventgpt}, and then apply the same human filtering procedure to obtain \textbf{RGBE-Chat}. Together, PEOD-Chat and RGBE-Chat serve as both training corpora and evaluation benchmarks; their sources and scales are summarized in \cref{tab:peod_rgbe_chat_bench}.

\begin{table}[hbtp]
\centering
\caption{Composition of PEOD-Chat and RGBE-Chat datasets. Data sources and post-screening sample counts for captioning and VQA tasks. In RGBE\text{-}ImageNet, the RGB data come from ImageNet~\cite{deng2009imagenet}, and the paired event data are generated using the method of~\cite{yang2025ezsr}.}
\small
\setlength{\tabcolsep}{8pt}
\renewcommand{\arraystretch}{1.15}
\begin{tabular}{l l r}
\toprule
\textbf{Dataset} & \textbf{Source} & \textbf{Size} \\
\midrule
\multirow{1}{*}{PEOD\text{-}Chat}  & PEOD~\cite{cui2025peodpixelalignedeventrgbbenchmark}              & 11k \\
\cmidrule(lr){2-3}
\rowcolor{gray!10}                 & \textbf{TOTAL}    & \textbf{11k} \\
\cmidrule(lr){1-3}
\addlinespace[0.5ex]
\multirow{6}{*}{RGBE\text{-}Chat}  & RGBE\text{-}ImageNet & 60k \\
                                   & DSEC~\cite{gehrig2021dsec}                & 12k \\
                                   & DDD17~\cite{binas2017ddd17}               & 1.7k \\
                                   & RGBE\text{-}SEG~\cite{chen2024segment}      & 15k \\
                                   & MVSEC~\cite{zhu2018multivehicle}               & 11k \\
                                   & M3ED~\cite{chaney2023m3ed}                & 14k \\
\cmidrule(lr){2-3}
\rowcolor{gray!10}                 & \textbf{TOTAL}      & \textbf{113.7k} \\
\bottomrule
\end{tabular}
\label{tab:peod_rgbe_chat_bench}
\end{table}

\section{Method}
\label{sec:method}
\begin{figure*}[hbtp]
    \centering
    \scalebox{1}[0.95]{\includegraphics[trim=1cm 18cm 15.5cm 1.6cm, clip, width=1\textwidth]{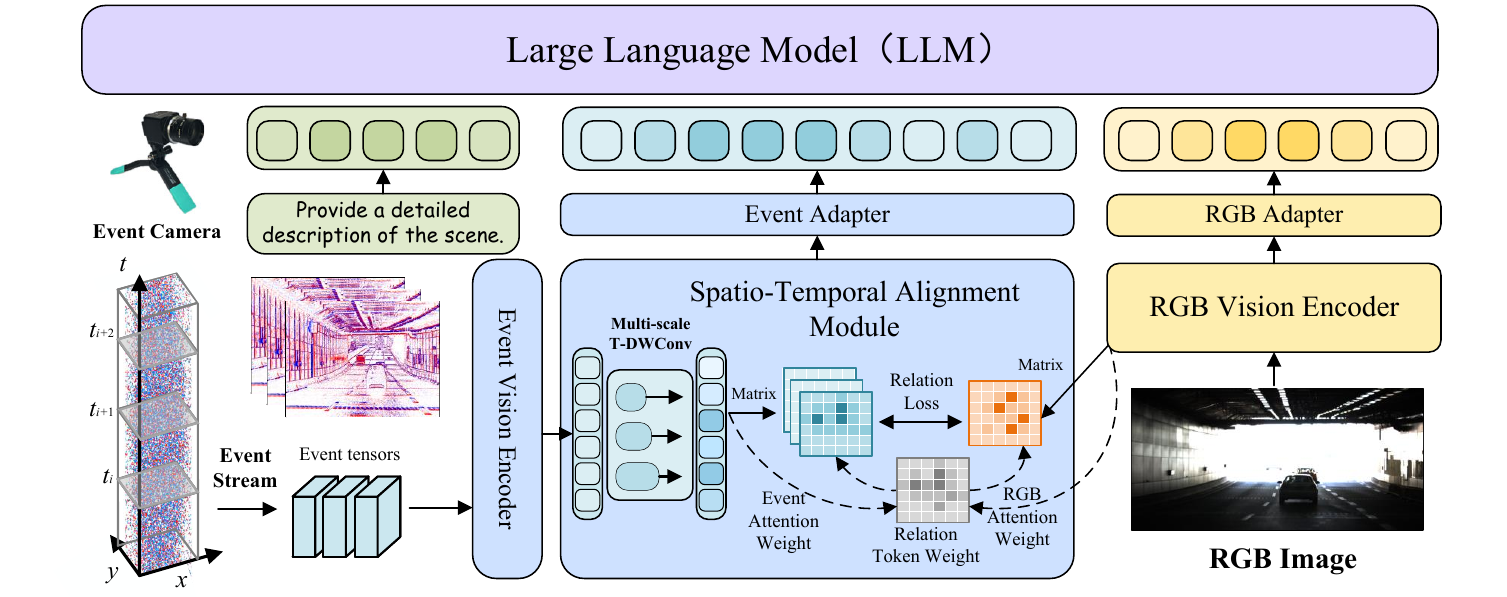}}
    \caption{RE-VLM model architecture. Synchronized RGB and event streams are encoded. During \emph{training}, a Spatio-Temporal Alignment Module (STAM) provides alignment signals and a relation loss. During inference, event features after the temporal DWConv are projected (event adapter) and, together with RGB tokens (RGB adapter), are fed to the LLM.}
    \label{fig:model_frame}
\end{figure*}
\subsection{Overview of RE-VLM}
\label{subsec:overview}
We aim to achieve robust scene understanding under adverse conditions by fusing fine-grained RGB appearance cues with event-driven dynamics. As shown in \cref{fig:model_frame}, RE-VLM leverages the event stream's high temporal resolution and high dynamic range together with the RGB image's static appearance, enabling comprehensive understanding in low-light or HDR scenes. The architecture comprises an RGB vision encoder, an event vision encoder, a lightweight Spatio-Temporal Alignment Module (STAM, \emph{training-time only}), two modality adapters, and an LLM decoder.

Given an RGB frame $X$ and its temporally aligned event stream $S$, the encoders produce feature maps
\begin{equation}
F_i = f_{\mathrm{rgb}}(X), \qquad F_e = f_{\mathrm{event}}(S).
\label{eq:enc-feats}
\end{equation}
A temporal depthwise convolution (DWConv) over the event slice axis yields an event representation $\tilde F^{\,e}$ used for projection, while STAM is used only during training to compute alignment signals and the relation loss. The modality adapters map features to the LLM space:
\begin{equation}
T_i = g_i\!\big(F_i\big), \qquad T_e = g_e\!\big(\tilde F^{\,e}\big).
\label{eq:adapters}
\end{equation}
At inference, the LLM receives instruction tokens $P$ concatenated with the two projected streams and performs causal decoding:
\begin{equation}
A = f_{\mathrm{LLM}}\!\big([P;\,T_i;\,T_e]\big).
\label{eq:llm}
\end{equation}

\subsection{Event Dynamics Encoding}
\label{subsec:event-dynamics}
Each raw event is represented as $e_j=(x_j,y_j,t_j,p_j)$ with polarity $p_j\in\{+1,-1\}$. We consider a short temporal window around the image timestamp and divide it into $N_w=3$ slices; events falling in slice $t$ are accumulated into a two-channel image $E_t$. Passing the sequence $\{E_t\}_{t=1}^{N_w}$ through an event encoder (a ViT backbone) produces per-slice feature maps $F^e_t = f_{\text{event}}(E_t)$ and thus a spatio\mbox{-}temporal tensor
\begin{equation}
F^e=\{F^e_t\}_{t=1}^{N_w}\in\mathbb{R}^{N_w\times H\times W\times D}.
\label{eq:event-tensor}
\end{equation}
To capture motion at multiple time scales, we apply multi\mbox{-}scale depthwise 1D convolutions along the temporal axis of $F^e$, concatenate their outputs, and project back to $D$ channels, yielding an enhanced event feature tensor $\widetilde{F}^e$. A lightweight SE\mbox{-}style temporal weighting is then applied to $\widetilde{F}^e$ to re\mbox{-}emphasize salient motion intervals and suppress background slices, before projecting the event features into the LLM space via $g_e(\cdot)$.

\noindent\textbf{Training-time Alignment with STAM.}
To encourage the RGB and event features to align in time and space, we introduce a Spatio\mbox{-}Temporal Alignment Module (STAM) during training. STAM computes two parallel self\mbox{-}attention mappings and then fuses them to identify important regions across both modalities. Concretely, we first resample the image feature map $F^i$ (from the RGB encoder at the same timestamp) and the event feature $\widetilde{F}^e$ to a shared spatio\mbox{-}temporal lattice of size $(T_c, H_c, W_c)$ for alignment. Let $\widetilde{R}^{(t)}, \widetilde{E}^{(t)} \in \mathbb{R}^{D\times H_c\times W_c}$ denote the aligned RGB and event feature grids at time step $t$. We flatten these feature maps into token matrices $R^{(t)}\in\mathbb{R}^{D\times N_r}$ afnd $E^{(t)}\in\mathbb{R}^{D\times N_e}$ (with $N_r=N_e=H_cW_c$). We then perform channel\mbox{-}wise $L_2$ normalization on each token vector to obtain $\widehat{R}^{(t)}=\operatorname{norm}_D(R^{(t)})$ and $\widehat{E}^{(t)}=\operatorname{norm}_D(E^{(t)})$. STAM's dual self\mbox{-}attention is computed by taking dot\mbox{-}products within each modality:
\begin{equation}
P_r^{(t)}=\widehat{R}^{(t)\top}\widehat{R}^{(t)},\qquad
P_e^{(t)}=\widehat{E}^{(t)\top}\widehat{E}^{(t)}.
\label{eq:stam_pr_pe}
\end{equation}
From each self\mbox{-}attention matrix, we derive a token saliency as the graph degree (row\mbox{-}sum of affinities): for modality $m\in\{r,e\}$, the degree vector from $P_m^{(t)}$ is reshaped to the spatial grid, yielding an importance map $\widetilde{w}_m^{(t)} \in \mathbb{R}^{H_c\times W_c}$. We fuse the two modalities' saliency maps by averaging and normalizing the result to avoid dominance by either modality, producing a unified importance map for each frame $t$:
\begin{equation}
w^{(t)}=\mathrm{norm}\!\left(\tfrac{1}{2}\big(\widetilde{w}_r^{(t)}+\widetilde{w}_e^{(t)}\big)\right).
\label{eq:fusew}
\end{equation}

\noindent\textbf{Relation Loss during Training.}
Using these fused importance weights, we define a relation loss to align the two modalities' features at the token level. First, we compute a per\mbox{-}frame discrepancy map $D^{(t)} \in \mathbb{R}^{H_c\times W_c}$ as the channel\mbox{-}wise mean absolute difference between the resampled RGB and event feature maps:
\begin{equation}
D^{(t)}_{h,w}=\frac{1}{D}\sum_{c=1}^{D}\big|\widetilde{R}^{(t)}_{c,h,w}-\widetilde{E}^{(t)}_{c,h,w}\big|.
\label{eq:disc}
\end{equation}
We then take the spatial inner product of the importance map with this discrepancy map, $\langle w^{(t)}, D^{(t)}\rangle$, which yields a single alignment penalty for frame $t$. The cross\mbox{-}modal alignment regularizer is defined as the average of these weighted discrepancies over all $T_c$ frames:
\begin{equation}
L_{\text{CA\mbox{-}WTD}}=\frac{1}{T_c}\sum_{t=1}^{T_c}\langle w^{(t)}, D^{(t)}\rangle.
\label{eq:cawtd}
\end{equation}
This loss penalizes feature mismatches between the two modalities, with a stronger effect on important regions (where $w^{(t)}$ is high), thereby pulling RGB and event representations closer in those areas. The total training objective adds this regularizer to the main VLM loss:
\begin{equation}
L = L_{\text{LLM}} + \lambda\,L_{\text{CA\mbox{-}WTD}},
\label{eq:total-loss}
\end{equation}
where $L_{\text{LLM}}$ is the standard language\mbox{-}guided loss and $\lambda$ controls the strength of the alignment. We set $\lambda = 0.1$ in our experiments, following the common practice of using a small weight to balance auxiliary alignment losses~\cite{lee2025multimodal}. 

\subsection{Training Pipeline}
\label{subsec:train}
\begin{figure}[hbtp]
    \centering
    \includegraphics[trim=0.7cm 12cm 9.2cm 0.5cm, clip, width=1\columnwidth]{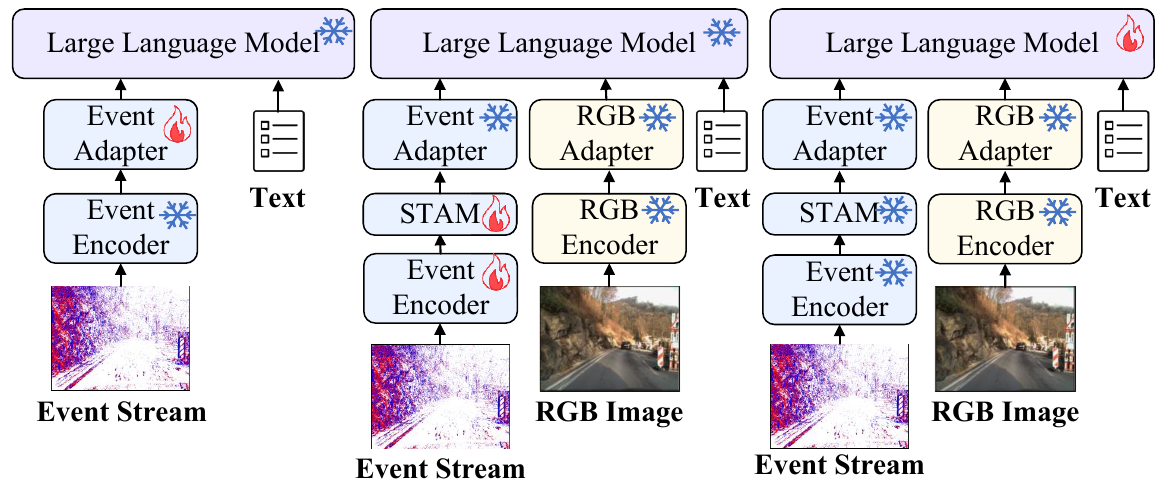}
    \caption{\textbf{Training pipeline.}
    Three compact stages: (1) Initial event--language alignment, (2) Align the event and RGB modalities with STAM, (3) End-to-end instruction tuning.
    }
    \label{fig:training}
\end{figure}
We adopt a concise three-stage curriculum that first aligns event representations with language, then aligns it with the RGB representation via STAM, and finally performs lightweight instruction tuning on the LLM. 

\noindent\textbf{Stage~1: Event-Language alignment.}
Starting from a pretrained RGB-based VLM backbone, we introduce the event vision encoder and event adapter and train them on a large-scale set of Event-Text caption pairs from RGBE-ImageNet dataset while keeping the LLM and the RGB branch frozen. This stage yields an event-only pathway whose representations are directly aligned with the language space.

\noindent\textbf{Stage~2: Event-RGB alignment.}
We then leverage paired RGB-Event data from ImageNet and N-ImageNet to align the event branch with the frozen RGB branch with captions serving as supervision. During this stage, the LLM and RGB pathway remain frozen; we optimize the event vision encoder together with the STAM module, which performs temporal modeling on event tokens and imposes relation losses between event and RGB features. This step builds a coherent dual-branch representation.

\noindent\textbf{Stage~3: Instruction tuning.}
Finally, we freeze both visual branches and STAM, and attach low-rank LoRA adapters to the LLM for supervised fine-tuning on multimodal instruction data (captioning and VQA dialogues). Only the LoRA parameters are updated, preserving the aligned RGB and event encoders while endowing \textbf{RE-VLM} with strong instruction-following ability and flexible inference under RGB-only, Event-only, or joint RGB+Event settings.

\section{Experiments}
\label{sec:exp}

\begin{table*}[hbtp]
\centering
\caption{Results on \textbf{PEOD-Chat} and \textbf{RGBE-Chat}. We report LLM-judge scores for Caption and VQA. Our RE-VLM attains the best performance on both benchmarks, with large gains on illumination-challenged PEOD-Chat, showing that jointly leveraging event streams and RGB improves illumination-challenged scene understanding. (*) denotes variants fine-tuned on PEOD-Chat and RGBE-Chat: the RGB-only model is trained with RGB--text supervision, and the event-only model with event--text supervision.}
\small
\setlength{\tabcolsep}{6pt}
\renewcommand{\arraystretch}{1.15}
\begin{tabular}{l l c *{10}{c}}
\toprule
\multirow{3}{*}{\textbf{Input}} & \multirow{3}{*}{\textbf{Models}} & \multirow{3}{*}{\textbf{Params}} & \multicolumn{5}{c}{\textbf{PEOD-Chat}} & \multicolumn{5}{c}{\textbf{RGBE-Chat}}\\
& & & \multicolumn{3}{c}{\textbf{Caption}} & \multicolumn{2}{c}{\textbf{VQA}} & \multicolumn{3}{c}{\textbf{Caption}} & \multicolumn{2}{c}{\textbf{VQA}}\\
\midrule
& & & \textbf{CI} & \textbf{DO} & \textbf{CU} & \textbf{Ave} & \textbf{Acc} & \textbf{CI} & \textbf{DO} & \textbf{CU} & \textbf{Ave} & \textbf{Acc} \\

\cmidrule(lr){4-6}\cmidrule(lr){7-8}\cmidrule(lr){9-11}\cmidrule(lr){12-13}
\multirow{6}{*}{\textbf{RGB-only}}
& Qwen2.5-VL~\cite{bai2025qwen2}        & 3B  & 2.47 & 2.03 & 3.04 & 3.47 & 0.52 & 3.34 & 2.70 & 3.64 & 3.80 & 0.66 \\

& Intern2VL~\cite{chen2024internvl}         & 4B  & 3.09 & 2.38 & 3.68 & 3.36 & 0.49 & 3.47 & 2.84 & 3.91 & 3.70 & 0.68 \\
& DeepSeek2-VL~\cite{lu2024deepseek}       & 7B  & 3.25 & 2.42 & 3.73 & 3.37 & 0.50 & 3.63 & 2.89 & 4.11 & 3.49 & 0.52 \\
& LLaVA-1.5~\cite{liu2023visual}       & 7B & 2.71 & 2.05 & 3.03 & 3.59 & 0.54 & 3.12 & 2.31 & 3.60 & 3.69 & 0.62 \\
& Qwen2.5-VL$^\ast$~\cite{bai2025qwen2} & 3B  & 3.23 & 2.74 & 3.51 & 3.61 & 0.55 & 3.91 & 3.41 & 4.27 & 3.86 & 0.65 \\
\midrule
\multirow{2}{*}{\textbf{Event-only}}
& EventGPT~\cite{liu2025eventgpt}          & 7B  & 2.51 & 2.06 & 2.65 & 3.04 & 0.40 & 2.82 & 2.34 & 3.08 & 3.10 & 0.39 \\
& Qwen2.5-VL$^\ast$~\cite{bai2025qwen2} & 3B  & 2.74 & 2.48 & 2.97 & 3.24 & 0.45 & 2.79 & 2.57 & 3.16 & 3.59 & 0.58 \\
\midrule
\rowcolor{gray!20}
\textbf{RGB+Event}
& \textbf{RE-VLM}   & 4B  & \textbf{3.68} & \textbf{3.12} & \textbf{3.95} & \textbf{3.82} & \textbf{0.63} & \textbf{4.03} & \textbf{3.50} & \textbf{4.34} & \textbf{4.20} & \textbf{0.75} \\
\bottomrule
\end{tabular}
\label{tab:REGPT_results_revised}
\end{table*}
\subsection{Experimental Settings}
\noindent\textbf{Datasets.} We construct two multimodal dialogue benchmarks tailored for challenging illumination and general scenes. From \mbox{PEOD-Chat}, we curate 1{,}750 test samples, with a distribution matching PEOD~\cite{cui2025peodpixelalignedeventrgbbenchmark}: 60\% adverse imaging (e.g., low light, overexposure, motion blur) and 40\% normal illumination. All text supervision from Section~\ref{sec:data_engineering} (captions and VQA pairs) manually corrected for factual accuracy and consistent references. From \mbox{RGBE-Chat}, we similarly curate 2{,}047 test samples covering diverse real-world scenes. We enforce a strict sequence-level split, ensuring none of the test sequences appear in the training set.

\noindent\textbf{Model \& Training.}
Our RE-VLM backbone is based on Qwen2.5-VL-3b.
Models are trained on $8\times$NVIDIA 4090 GPUs, proceeding in three stages:
(i) \emph{Event-Language alignment} on \textbf{1{,}300K} pairs with learning rate $1\times10^{-4}$ and batch size $32$;
(ii) \emph{Event-RGB alignment} on \textbf{600K} pairs with learning rate $1\times10^{-4}$ and batch size $16$;
(iii) \emph{Instruction tuning} on \textbf{120K} samples with learning rate $2\times10^{-4}$ and batch size $16$.
To retain an end-to-end pipeline and ensure fair comparison under identical inputs, event streams are rendered into event images for all competing models.

\noindent\textbf{Evaluation Protocol and Metrics.}
Following the LLM-as-a-judge protocol popularized by Video-ChatGPT~\cite{maaz2024video}, we conduct \emph{zero-shot} evaluation on VQA-style prompts. \textbf{GPT-3.5-Turbo} acts as the judge and assigns \mbox{0--5} Likert scores along three dimensions:
\textbf{CI} (Correctness of Information), \textbf{DO} (Detail Orientation), and \textbf{CU} (Contextual Understanding). For the VQA responses, we compute two overall measures: \textbf{Ave}, the LLM-assigned average VQA answer quality score (0--5 scale), and \textbf{Acc}, a VQA accuracy metric computed at the attribute level (awarding 1 point for each correctly predicted attribute, and 0 for any incorrect attribute).
\subsection{Comparison with State-of-the-Art Models}
\paragraph{Quantitative Results.}
\cref{tab:REGPT_results_revised} compares our RE-VLM to recent MLLMs and an event-only VLM. Across both benchmarks, \textbf{RE-VLM} achieves the highest scores on captioning metrics (CI, DO, CU) and VQA performance (Ave, Acc), indicating that jointly leveraging event streams alongside RGB images significantly improves a model's understanding of complex scenes. The gains are particularly pronounced on the illumination-challenged PEOD-Chat, where RGB image quality is degraded; in these cases, RE-VLM recovers motion and structural cues from the event data while providing color and texture information via the RGB branch that event-only models inherently lack. On the more diverse RGBE-Chat benchmark, RE-VLM still yields improvements over all baselines, suggesting that the proposed dual-stream design generalizes beyond strictly adverse conditions.

\begin{figure}[hbtp]
    \centering
    \includegraphics[trim=2cm 0.58cm 7.5cm 1.7cm, clip, width=1\columnwidth]{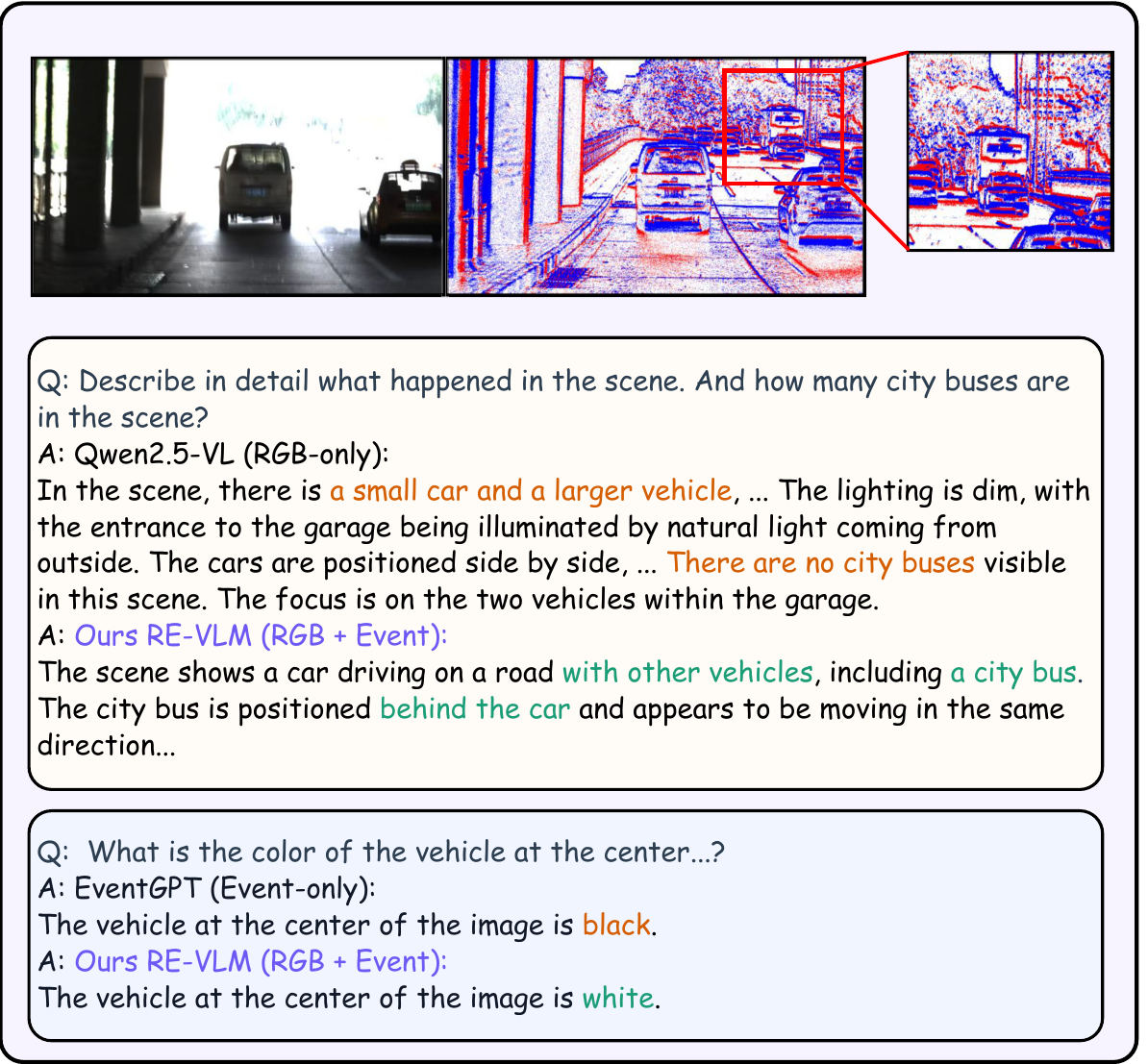}
    \caption{Qualitative VQA comparison in an overexposed traffic scene. RGB-only and event-only baselines miss the city bus or fail to capture the color, while RE-VLM correctly identifies both, demonstrating robust scene understanding under challenging lighting conditions.
    }
    \label{fig:Qualitative}
\end{figure}

\noindent\textbf{Qualitative analysis.} \cref{fig:Qualitative} presents a qualitative comparison on a challenging scene. The RGB-only baseline (Qwen2.5-VL) fails to detect a city bus under severe overexposure in the scene, erroneously reporting no bus present, whereas RE-VLM leverages motion cues from the event stream to correctly identify a city bus behind the car. Similarly, the event-only model (EventGPT) cannot determine appearance attributes such as color while RE-VLM accurately recognizes it as white by integrating the RGB image's color and texture cues. These results highlight that by fusing complementary motion and appearance information, RE-VLM can robustly identify object categories and attributes even under adverse conditions where single-modality models falter.

\subsection{Ablation Study}
\paragraph{Single-branch Inference and STAM Module.}
RE-VLM can operate not only with joint RGB+Event input, but also with a single modality at inference time. \cref{tab:ablation-peod} and \cref{tab:ablation-rgbe} present an ablation study on input modalities and our Spatio-Temporal Alignment Module (STAM). 
On both benchmarks, using only RGB or only event data still yields reasonable captioning and VQA performance. The joint RGB+Event configuration consistently outperforms either single-modality setting, demonstrating that our dual-branch architecture effectively exploits the complementary information from events and RGB, especially under adverse conditions. We also examine the contribution of the STAM module by comparing against a variant that replaces STAM with naive feature concatenation. Incorporating STAM leads to consistent improvements or maintenance of performance across both datasets.

\begin{table}[hbtp]
\centering
\caption{Ablation on input modality and STAM on \textbf{PEOD-Chat}. Single-branch rows drop the other stream at inference. For RGB+Event, \xmark\ indicates that STAM is not used. }
\small
\setlength{\tabcolsep}{6pt}
\renewcommand{\arraystretch}{1.10}
\begin{tabular}{l c *{5}{c}}
\toprule
\multirow{2}{*}{\textbf{Input}} & \multirow{2}{*}{\textbf{STAM}} &
\multicolumn{3}{c}{\textbf{Caption}} & \multicolumn{2}{c}{\textbf{VQA}}\\
 & & \textbf{CI} & \textbf{DO} & \textbf{CU} & \textbf{Ave} & \textbf{Acc} \\
\cmidrule(lr){1-2}\cmidrule(lr){3-5}\cmidrule(lr){6-7}
RGB-only    & \textemdash & 3.05 & 2.51 & 3.32 & 3.63 & 0.57 \\
Event-only  & \textemdash & 2.82 & 2.57 & 3.09 & 3.40 & 0.48 \\
RGB+Event   & \xmark          & 3.62   & 3.08   & 3.91   & 3.79   & 0.61   \\
RGB+Event   & \cmark         & 3.68 & 3.12 & 3.95 & 3.82 & 0.63 \\
\bottomrule
\end{tabular}
\label{tab:ablation-peod}
\end{table}
\begin{table}[hbtp]
\centering
\caption{Ablation on input modality and STAM on \textbf{RGBE-Chat}. Single-branch rows drop the other stream at inference. For RGB+Event, \xmark\ indicates that STAM is not used. }
\small
\setlength{\tabcolsep}{6pt}
\renewcommand{\arraystretch}{1.10}
\begin{tabular}{l c *{5}{c}}
\toprule
\multirow{2}{*}{\textbf{Input}} & \multirow{2}{*}{\textbf{STAM}} &
\multicolumn{3}{c}{\textbf{Caption}} & \multicolumn{2}{c}{\textbf{VQA}}\\
 & & \textbf{CI} & \textbf{DO} & \textbf{CU} & \textbf{Ave} & \textbf{Acc} \\
\cmidrule(lr){1-2}\cmidrule(lr){3-5}\cmidrule(lr){6-7}
RGB-only    & \textemdash & 3.97 & 3.46 & 4.32 & 4.10 & 0.73 \\
Event-only  & \textemdash & 2.49 & 2.48 & 2.82 & 3.53 & 0.57 \\
RGB+Event   & \xmark          & 4.01   & 3.47   & 4.35   & 4.19   & 0.74   \\
RGB+Event   & \cmark         & 4.03 & 3.50 & 4.34 & 4.20 & 0.75 \\
\bottomrule
\end{tabular}
\label{tab:ablation-rgbe}
\end{table}

\noindent\textbf{Cross-Validation with an Open-Source LLM Judge.}
Because GPT-3.5-Turbo is closed-source, and following recent recommendations to cross-validate LLM-as-a-judge evaluations~\cite{zheng2023judging}, we additionally employ the open-source \textbf{Qwen3-Omni-30B}~\cite{xu2025qwen3} as an independent judge on PEOD-Chat. \cref{tab:eval_gptturbo} shows consistent trends across metrics, corroborating the conclusions drawn with GPT-3.5-Turbo.

\begin{table}[t]
\centering
\caption{Evaluation using an open-source LLM judge (Qwen3-Omni-30B~\cite{xu2025qwen3}) in place of GPT-3.5-Turbo. The trends remain consistent with the GPT-3.5-Turbo evaluation across metrics.}
\label{tab:eval_gptturbo}
\small
\setlength{\tabcolsep}{5pt}
\renewcommand{\arraystretch}{1.15}
\begin{tabular}{lcccc}
\toprule
\textbf{Task} & \textbf{Metric} & \textbf{Qwen2.5VL} & \textbf{EventGPT} & \textbf{RE-VLM} \\
\midrule
\multirow{3}{*}{Caption} 
& \textbf{CI} & 2.17 & 1.99 & 3.29 \\
& \textbf{DO} & 1.82 & 1.89 & 3.45 \\
& \textbf{CU} & 2.71 & 2.50 & 3.85 \\
\cmidrule(lr){1-5}
\multirow{2}{*}{VQA}
& \textbf{Ave} & 2.78 & 2.38 & 3.43 \\
& \textbf{Acc} & 0.48 & 0.40 & 0.62 \\
\bottomrule
\end{tabular}
\end{table}

\section{Conclusion}
We introduced RE-VLM, a dual-stream vision--language model that jointly leverages RGB images and event streams for robust scene understanding. To address the scarcity of RGB-Event-Text supervision, we further introduced a graph-driven data generation pipeline and constructed two datasets covering both common and challenging conditions. Experiments show that RE-VLM consistently outperforms strong RGB-only and event-only baselines, particularly in challenging scenarios, providing a practical foundation for future work on event-augmented vision-language understanding.

\clearpage
\section*{Acknowledgments}
This work was supported by the National Key R\&D Program of China (2021ZD0109802), the Beijing Natural Science Foundation (4262060), the BUPT Innovation and Entrepreneurship Support Program (2025-YC-T026), the Opening Project of the State Key Laboratory of General Artificial Intelligence, BIGAI/Peking University (Project No.\ SKLAGI2025OP23), and the High-performance Computing Platform of BUPT.

{\small
\bibliographystyle{ieeenat_fullname}
\bibliography{11_references}
}

\end{document}